\documentclass[letterpaper, 10 pt, conference]{ieeeconf}  % Comment this line out if you need a4paper

\IEEEoverridecommandlockouts                              % This command is only needed if 
\title{\LARGE \bf
Ground-level Viewpoint Vision-and-Language Navigation in \\ Continuous Environments
}

\author{Zerui Li, Gengze Zhou, Haodong Hong, Yanyan Shao, Wenqi Lyu, Yanyuan Qiao and Qi Wu*% <-this % stops a space
\thanks{Zerui Li, Gengze Zhou, Wenqi Lyu, Yanyuan Qiao and Qi Wu are with the Australian Institute for Machine Learning, The University of Adelaide, Adelaide 5005, Australia. Haodong Hong is with The University of Queensland, Brisbane 4072, Australia. Yanyan Shao is with Zhejiang University of Technology, Hangzhou 310014, China}%
\thanks{*Corresponding Author: Qi Wu  ({\tt\footnotesize qi.wu01@adelaide.edu.au})}% <-this % stops a space
}
\usepackage[dvipsnames]{xcolor}
\usepackage{gensymb}
\usepackage{colortbl}
\usepackage{multirow}
\usepackage{amsmath}
\usepackage{amssymb}
\usepackage{graphicx}
\usepackage{ulem}

\usepackage{textcomp}

\usepackage{booktabs}
\usepackage{bbding}
\usepackage{pifont}
\usepackage{wasysym}
\newcommand{\cmark}{\ding{51}}%
\newcommand{\xmark}{\ding{55}}%

\begin{document}

\maketitle
\thispagestyle{empty}
\pagestyle{empty}

%%%%%%%%%%%%%%%%%%%%%%%%%%%%%%%%%%%%%%%%%%%%%%%%%%%%%%%%%%%%%%%%%%%%%%%%%%%%%%%%
\vspace{-4pt}
\begin{abstract}

Vision-and-Language Navigation (VLN) empowers agents to associate time-sequenced visual observations with corresponding instructions to make sequential decisions. However, generalization remains a persistent challenge, particularly when dealing with visually diverse scenes or transitioning from simulated environments to real-world deployment. In this paper, we address the mismatch between human-centric instructions and quadruped robots with a low-height field of view, proposing a Ground-level Viewpoint Navigation (GVNav) approach to mitigate this issue. This work represents the first attempt to highlight the generalization gap in VLN across varying heights of visual observation in realistic robot deployments.
Our approach leverages weighted historical observations as enriched spatiotemporal contexts for instruction following, effectively managing feature collisions within cells by assigning appropriate weights to identical features across different viewpoints. This enables low-height robots to overcome challenges such as visual obstructions and perceptual mismatches. Additionally, we transfer the connectivity graph from the HM3D and Gibson datasets as an extra resource to enhance spatial priors and a more comprehensive representation of real-world scenarios, leading to improved performance and generalizability of the waypoint predictor in real-world environments. Extensive experiments demonstrate that our Ground-level Viewpoint Navigation (GVnav) approach significantly improves performance in both simulated environments and real-world deployments with quadruped robots.

\end{abstract}
\vspace{-5pt}

%%%%%%%%%%%%%%%%%%%%%%%%%%%%%%%%%%%%%%%%%%%%%%%%%%%%%%%%%%%%%%%%%%%%%%%%%%%%%%%%

\section{INTRODUCTION}
\vspace{-3pt}
Vision-and-Language Navigation (VLN) is a challenging cross-domain research field that requires an agent to interpret natural language instructions from humans and navigate in unseen environments by executing a sequence of actions. 

There have been significant advancements in understanding and aligning vision, language, and action in navigation tasks~\cite{anderson2018vision,qi2020reverie, ku2020room, hong2022bridging, wang2023scaling}, nevertheless, the effectiveness of these developments is limited when applied to practical scenarios, as they are primarily designed in discrete environments, where the agent can only navigate on predefined navigation graph by teleporting between adjacent nodes. Therefore, Krantz et al.~\cite{krantz2020beyond} proposed a benchmark that sets the VLN task in a continuous photo-realistic reconstructing 3D environment where visual agents are required to execute low-level discrete actions.
Irshad et al.\cite{irshad2021hierarchical} introduced a hierarchical model to better simulate real robotic actions by estimating the agent's linear and angular velocities as continuous actions within the Robo-VLN environment. Recently, based on the close-to-human~\cite{anderson2018vision} performance of VLN tasks both in discrete settings~\cite{wang2023scaling} (over 80\% successful rate) and continuous settings~\cite{an2022bevbert} (over 60\% successful rate), researchers are extending the VLN task into real robot experiments ~\cite{zhang2024navid, wang2024sim, li2024human, yokoyama2024vlfm}. However, a significant performance gap between simulation and real-world deployment has been identified.

One of the primary reasons for this gap is the mismatch of panoramic observation in VLN research and monocular observation on real robots. Most existing Sim-to-Real VLN models rely on monocular RGBD cameras as visual sensors, limiting the agent’s field of view and preventing panoramic observation. This restricted visual input hinders the agent’s ability to perceive the environment and make informed decisions. Zhang et al.~\cite{zhang2024vision} demonstrated that panoramic visual input significantly outperforms monocular input across various performance metrics, further emphasizing the limitations of monocular sensors in real-world applications.

%形象的例子（人视角狗视角区别）
\begin{figure}[t]
      \centering
      \includegraphics[width=.99\linewidth]{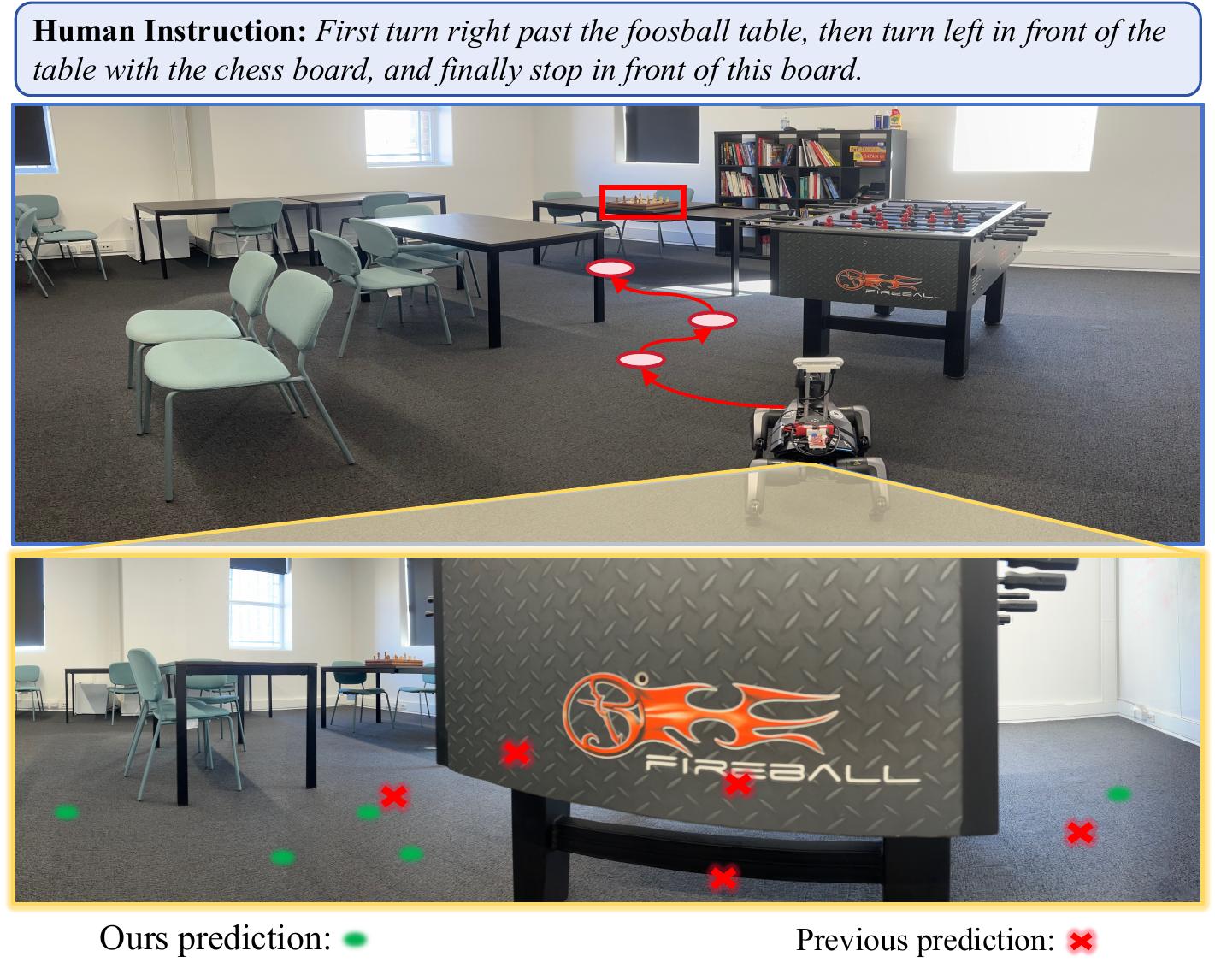}
      \vspace{-25pt}
      \caption{There is a significant viewpoint height discrepancy between humans and the robot dog (Up: human, Down: dog). Humans typically have a much higher line of sight compared to the robot dog. Our waypoint prediction network could provide robust prediction under a low line of sight.}
      \label{view}
\vspace{-20pt}
\end{figure}

Moreover, in real-world applications where humans issue commands and robots execute actions, there is often a significant discrepancy in viewpoint height between humans and most robots, such as a robot dog. Humans typically have a much higher line of sight, allowing them to observe a broader and more comprehensive view of the environment. In contrast, the robot dog’s lower viewpoint limits its field of vision, focusing more on ground-level obstacles and localized surroundings. This height disparity introduces an information asymmetry: humans issue commands based on a global understanding of the environment, while the robot dog, constrained by its limited perspective, makes decisions based on partial, localized information. This mismatch can lead to errors in command interpretation, particularly in complex environments where the robot lacks sufficient information to execute tasks accurately such as shown in Figure~\ref{view}, and this mismatch can not be solved by simply raise the height of the dog's sensor since it will decrease the passability in constrained environments, potentially impeding its ability to maneuver through narrow spaces or under obstacles.

To the best of our knowledge, current VLN research has not adequately addressed the impact of this visual information gap on performance, which poses practical challenges in applications involving various forms, such as assistive robots autonomous vehicles. Exploring these gaps is crucial for improving VLN tasks in real-world robots, which vary in shape and visual perspective. In this paper, we identify several challenges in deploying VLN systems on real robots, using the Xiaomi Cyberdog—a typical small dog-shaped robot with a low line of sight—as a case study:
(1) VLN methods are navigating through panoramic observation, but most of the robots are constructed with monocular RGBD cameras as visual sensors. 
(2) There are numerous visual domain variances to transfer the VLN model from a simulator to the real world:
Firstly, small-sized dog-like robots such as Unitree Go1 and Xiaomi Cyberdog are only around 30cm in height. The reduction in the height of the viewpoint leads to a different understanding of landmarks.
Secondly, ground-level viewpoint also results in a significant performance drop in depth-only waypoint prediction used by most VLN-CE approaches~\cite{an2022bevbert, wang2023gridmm, an2023etpnav}.
Thirdly, instructions in existing datasets are primarily designed based on human's line of sight, which are not always suitable for quadruped robots.
(3) The generalizability of waypoint prediction has been underestimated in VLN-CE R2R benchmarks. The waypoint prediction does not perform well in more complex real environments.
%In this paper, we mainly focus on those three gaps. 
Our contribution includes:
\begin{enumerate}
\item We assessed the impacts of notable differences in visual information between human-issued instructions and robot dogs' execution by reconstructing the Xiaomi Cyberdog with a programmable motor to spin an RGBD camera to get panoramic visual input.
\item We assessed the impact of waypoint prediction on ground-level viewpoint between depth-only and RGBD waypoint predictions. We further transfer the connectivity graphs from public 3D scans as extra data to power up the generalization ability of waypoint predictors in real-world complex environments. 
\item We proposed an adaptive information-gathering module to handle obstruction in local observation by assigning appropriate weights to identical features across different viewpoints, significantly enhancing performance in both simulated environments and real-world deployments with quadruped robots.
\end{enumerate}
\section{RELATED WORK}

\subsection{Vision-and-Language Navigation}

In recent years, significant efforts have been devoted to enabling navigation in previously unvisited environments based on human instructions. This research is often conducted within discretized simulated scenes that utilize predefined navigation graphs~\cite{anderson2018vision, anderson2020rxr, qi2020reverie, thomason2020cvdn}. To facilitate the alignment of language and visual cues for decision-making, Fried et al.~\cite{fried2018speaker} introduced the concept of navigation through panoramic actions. This method allows the agent to teleport between adjacent nodes on the graph by selecting an image oriented toward the target node. Building on this foundation, research in VLN has made steady progress in improving model performance towards human-level capabilities~\cite{ma2019self, wang2019reinforced, fried2018speaker, tan2019envdrop, ke2019tactical, fu2020counterfactual, qi2020object, hong2020graph, hao2020towards,li2019robust,hong2020recurrent,majumdar2020improving,chen2022think,wang2023gridmm}. Recently, Wang et al.~\cite{wang2023scaling} achieved an 80\% single-run success rate on the widely recognized R2R-VLN benchmark~\cite{anderson2018r2r}. However, these advancements remain constrained by the limitations of the high-level panoramic action space when applied to real-world scenarios. To address this, Krantz et al.~\cite{krantz2020beyond} proposed a benchmark that shifts the VLN task from a discrete to a continuous environment, more closely resembling real-world settings. Despite this shift, directly transferring VLN methods into continuous environments has resulted in substantial performance declines~\cite{irshad2021hierarchical, irshad2022semantically, raychaudhuri2021language}. To overcome these challenges, several studies~\cite{hong2022bridging, krantz2020navgraph, krantz2021waypoint} have introduced waypoint models that bridge the gap between VLN and VLN-CE, maintaining the simplicity of learning cross-modal alignment in discrete environments. Notably, Hong et al.~\cite{hong2022bridging} highlight that the choice of waypoint direction and step size significantly impacts VLN policy decision-making. In this work, we aim to optimize waypoint prediction under conditions of limited line-of-sight.

\subsection{Vision-and-Language Navigation in Real Environments}

Recently, researchers have been trying to extend the VLN task in real robots. Navid~\cite{zhang2024navid} proposed a video-based large vision language model (VLM), it only requires an on-the-fly video stream from a monocular RGB camera equipped on the robot to output the next-step action with human instructions, to showcase the capability of VLMs to achieve state-of-the-art level navigation performance without any maps, odometers, or depth inputs. Wang et al.~\cite{wang2024sim} propose an approach to endow the monocular robots with panoramic traversability perception and panoramic semantic understanding.  This method transfers the high-performance panoramic VLN models to the common monocular robots and tested in real robots. Li et al.~\cite{li2024human} extended traditional VLN by incorporating dynamic human activities and relaxing key assumptions, and introduced a Human-Aware 3D (HA3D) simulator and also tested in a real robot. However, none of them indicates the performance dropping by generalization gap in different height field of view.

\section{PRELIMINARIES}

\subsection{VLN Background}

% Given an environment that can be represented as a graph \(G=(V,E)\), where \(V\) is the nodes representing different locations and \(E\) is the edges representing the navigable path between nodes. Each node \(v \in V\) corresponds to a physical location and has visual observations \( o_t\) associated with it at time \(t\), where \( o_t\) has RGBD images \( o_t^{\text{rgb}} \in \mathbb{R}^{H \times W \times 3} \) and \( o_t^{\text{depth}} \in \mathbb{R}^{H \times W} \). The agent receives a natural language instruction \( L = \{ l_1, l_2, \dots, l_n \} \), where \( l_i \) are tokens (words) in the instruction. The instruction tells the agent how to navigate from the starting location \( v_{\text{start}} \) to the goal location \( v_{\text{goal}} \).
As the navigation graph assumption cannot reflect the challenges a deployed system would experience in a real world environment. This paper focuses on the VLN-CE, an agent tasked with navigating through a continuous 3D environment based on natural language instructions. The environment represented as a continuous 3D spacese \(E\), where the agent’s position at any time \( t \) is given by its 3D coordinates \( \mathbf{x}_t = (x_t, y_t, z_t) \in \mathcal{E} \), where \( x_t \), \( y_t \), and \( z_t \) represent the agent’s location in a continuous space.
At each position \( \mathbf{x}_t \), the agent perceives its surroundings through visual observations \( o_t\), where \( o_t\) has RGBD images \( o_t^{\text{rgb}} \in \mathbb{R}^{H \times W \times 3} \) and \( o_t^{\text{depth}} \in \mathbb{R}^{H \times W} \).The agent is provided with a natural language instruction \( L = \{ l_1, l_2, \dots, l_n \} \), where \( l_i \) are tokens (words) in the instruction. This instruction guides the agent from a start position \( \mathbf{x}_{\text{start}} \in \mathcal{E} \) to the goal position \( \mathbf{x}_{\text{goal}} \in \mathcal{E} \) with discrete low-level actions.

\subsection{Cross-modal Planning with Topological Map}
\noindent\textbf{Waypoint Prediction Network:} Let \( \mathcal{P}_t = \{p_1, p_2, \dots, p_n\} \) represent the 3D waypoint positions at time step \( t \), where each \( p_i \in \mathbb{R}^3 \). Similarly, let \( \mathcal{V}_t = \{v_1, v_2, \dots, v_n\} \) denote the corresponding \( d \)-dimensional visual features. At each time step, the visual encoders process the panoramic input to generate \( \mathcal{V}_t \), and a Transformer operates on \( \mathcal{V}_t \) to establish spatial and contextual relationships among the neighboring sectors, enriching the visual feature representation and informing the generation of candidate waypoints \( \mathcal{P}_t \), where each waypoint is associated with a direction encoded in \( v_i \). The agent selects the most promising waypoint \( p_i \) based on its visual feature and spatial position, simplifying navigation by moving directly toward the chosen waypoint.

\noindent\textbf{Topological Navigation Policy:} To enable effective backtracking and planning in the continuous environment, we follow the previous SoTA method ETPNav~\cite{an2024etpnav} on VLN-CE and perform language-guided navigation based on topological mapping. The environment is represented as a graph-based topo map $G_t = \{N_t, E_t\}$ keeps track of all observed nodes along the path $\Gamma'$. Given $\Gamma'$, we initialize $G_t$ by deriving its corresponding sub-graph from the predefined graph $G^*$. The nodes $N_t$ are divided into three categories: 

\begin{itemize}
    \item \textbf{Visited Node} is the agent has already visited
    \item \textbf{Current Node} is where the agent is currently located
    \item \textbf{Ghost Node} is a hypothetical node representing an uncertain or predicted location in the environment, not yet confirmed
    % \item \textbf{Waypoints} are predicted possibly accessible locations near the agent
    
\end{itemize}

The edges $E_t$ record the Euclidean distances among all adjacent nodes. The feature vectors $V^p_t$ are mapped onto the nodes as their visual representations. Taking time step $t$ as an example, $V^p_t$ are first fed into a panoramic encoder~\cite{an2023etpnav} to obtain contextual view embeddings $\hat{V}^p_t$. \textbf{Visited Node} and \textbf{Current Node} have been visited and can access panoramas, they are represented by an average of panoramic view embeddings. \textbf{Ghost Node} is partially observed and therefore is represented by accumulated embeddings of views from which \textbf{Ghost Node} can be observed. $G_t$ is equiped with a global action space $A^G$ for long-term planning, which consists of all observed nodes.

The graph $G^*$ is updated continuously based on the agent’s predictions and spatial relationships between nodes. If a visited node is localized, the input waypoint is deleted, and an edge is added between the current node and the localized visited node. If a ghost node is localized, the position and visual representation of the input waypoint are accumulated to the localized ghost node. This means that the ghost node’s position and features are updated based on the accumulated observations of the waypoint. If no node is localized, the input waypoint is added to the graph as a new ghost node. This newly added ghost node will remain unconfirmed until future localization attempts.
To ensure the graph $G^*$ remains efficient, nodes that are too close together or redundant are pruned. If the distance between nodes \( v_i \) and \( v_j \) is less than a threshold \( \epsilon \), then 
prune \(v_i\) \text{ if } \( d(v_i, v_j) \) < \( \epsilon \), where \( d(v_i, v_j) \) is a distance function between two nodes.

%%%%%%%%%%%%%%%%%%%%%%%%%%%%%%%%%

\section{METHODS}

\begin{figure*}[t]
      \centering
      \includegraphics[width=.99\linewidth]{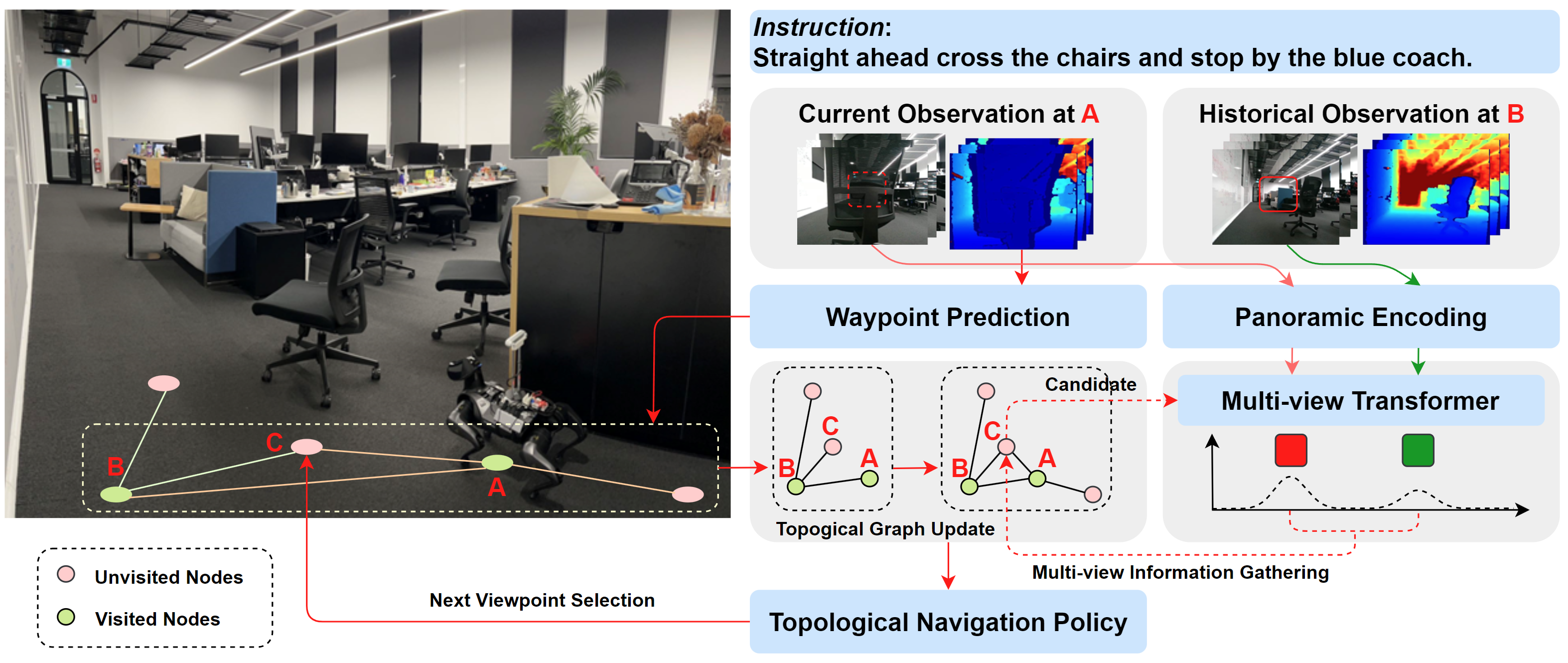}
      \vspace{-10pt}
      \caption{Multi-view Information Gathering emphasizes more informative features for the current context, enabling adaptive selection of the visual representations from multiple viewpoints (\textbf{A} and \textbf{B}). The navigation policy identifies the optimal next viewpoint in the topological graph (selecting \textbf{C} as the next viewpoint after \textbf{A}). This prediction is based not only on the robot's current observation at \textbf{A}, but also on previous, unobstructed views (from \textbf{B}), allowing the robot to mitigate occlusions and plan more robust navigation strategies.}
      \label{teaser}
    \vspace{-15pt}
\end{figure*}

\subsection{Scaling up Waypoint Prediction Network Training}

The first challenge posed by the ground-level viewpoint is the substantial degradation in waypoint prediction performance. This is not only due to the downward-shifted line of sight, which limits the visual field, but also the inherently low generalizability of the waypoint predictor in complex, real-world environments. Fig.~\ref{teaser} illustrates the candidate waypoints predicted by a waypoint predictor re-trained exclusively on the R2R dataset using low line-of-sight visual inputs, denoted by red crosses. Despite being re-trained to account for the robot's lower viewpoint, the predictions exhibit suboptimal performance. To solve this problem, we follow ScaleVLN~\cite{wang2023scaling} and construct a large waypoint prediction dataset in 800 scans from HM3D~\cite{ramakrishnan2021habitat}, 491 scans from Gibson~\cite{xia2018gibson}, and 61 scans from MP3D~\cite{chang2017matterport3d} under low-angle observation. Despite heuristically sampled viewpoints can also estimate feasible navigation paths based on depth information, it often lack the flexibility to handle complex, language-guided navigation tasks and oversimplify the navigation process by focusing primarily on depth cues, neglecting the rich semantic and visual information.

Specifically, we adopt the connectivity graph constructed in ScaleVLN~\cite{wang2023scaling} and discretize the environments into undirected graphs. At each node of the graphs, we annotate the distance and orientation of the connected nodes as ground truth supervision for the Waypoint Prediction Network. This provides 212924 training samples in total. Compared to the original training data for the waypoint prediction network~\cite{hong2022bridging}, this raised $\times 22.02$ in training data amount. Moreover, we set the rendering height to 80 cm from the ground in the Habitat simulator, and captured the depth images from ground-level observation at each node. 

\subsection{Multi-view Information Gathering}

The second challenge posed by the ground-level viewpoint is the discrepancy between the oracle and the agent's local observation, caused by environmental obstructions. This creates difficulties for the agent when attempting to predict the next action based on limited local observations. 
%Due to the restricted field of view at ground level, there is often a significant mismatch between the oracle, which represents a global, unobstructed viewpoint, and the robot's local observations, which can be occluded by objects in the environment. This constraint complicates the agent’s ability to accurately predict its next action based purely on its current state. For example, obstacles like furniture at lower viewpoints may obscure parts of the surrounding environment, leading to incomplete observations. 
In Figure~\ref{teaser}, the robot’s observation at position \textbf{A} is limited, while historical data from position \textbf{B} offers an unobstructed view. This disparity between local observations and the oracle’s ideal perspective becomes critical when the agent is tasked with selecting the next action or viewpoint in its decision-making.

%Local observations provide no meaningful information about the environment and can be considered highly noisy, the previous method that averaged these incomplete/noisy features with relevant, clear features from previous viewpoints dilutes the useful information. To overcome this problem, 
We propose to adaptively gather information from previous unobstructed angles along the trajectory from the previous SoTA method ETPNav~\cite{an2024etpnav} on VLN-CE. As illustrated in Figure~\ref{teaser}, during the update of the topo map with the predicted waypoints, we introduce a trainable transformer encoder layer that adaptively selects the optimal visual representation $\tilde{v}_g$ for each ghost node g. At each time step t, the visual representations $\mathcal{V^p}_t = \{v^p_1, v^p_2, \dots, v^p_n\}$ are processed through the trainable transformer encoder layer, which applies self-attention to capture dependencies between the visual features:
\[
V'_t = \text{SelfAttn}(V^p_t)
\]
Here, $V'_t \in \mathbb{R}^{n \times d}$ is the output matrix of the transformer, which incorporates the contextual relationships between the visual features. Instead of averaging the visual features, the transformer encoder layer uses a learned attention mechanism. Specifically, after applying the transformer block, the layer computes a set of learned weights $\mathcal{W} = \{w_1, w_2, w_e\}$for each input feature:

This generates attention weights, $W \in \mathbb{R}^{n \times 1}$, used to select the most relevant feature representations.
The final representation for the ghost node is computed as a weighted sum of the transformed features $v'_i$, where the weights are derived from the attention mechanism:
\[
\tilde{v}_g = \sum_{i=1}^{n} \text{Softmax}(\text{Linear}(v'_i)) v'_i
\]
Thus, the transformer encoder layer hence then learns to emphasize more informative features for the current context, enabling adaptive selection of the visual representations from multiple viewpoints (\textbf{A} and \textbf{B}). As shown in Figure~\ref{teaser}, the navigation policy identifies the optimal next viewpoint in the topological graph (selecting \textbf{C} as the next viewpoint after \textbf{A}). This prediction is based not only on the robot's current observation at \textbf{A}, but also on previous, unobstructed views (from \textbf{B}), allowing the robot to mitigate occlusions and plan more robust navigation strategies.

\section{EXPERIMENTS}
\vspace{-5pt}
\subsection{Experiment Setup}

\begin{table*}[t]
    \footnotesize
    \caption{
    The performance gap in VLN models trained on high line-of-sight perspective data that testing with low line-of-sight. \\
    % H $\rightarrow$ High Line-of-sight as default settings; 
    }
    % The best results are in \textbf{bold} font and the second ones are \underline{underlined}.}
    \label{tab:policy}
    \vspace{-15pt}
    \begin{center}
    % \begin{tabular}{llccccccc}
    \begin{tabular}{ccccccccccccccc}
    \hline
    \toprule
     % &  & & \multirow{12}{*}{Changing in Viewpoint} 
    % \multirow{2}{*}{\#ID} &\multirow{2}{*}{Method} &  \multicolumn{3}{c}{Mask}   &  \multicolumn{3}{c}{Text} &
    % \multirow{2}*{Overall} \\
    % \multirow{2}*{$\mathcal{J\&F}^*$}  \\ 

 % \hline
  % && &  \multicolumn{12}{c}{Changing in Viewpoint} \\

\multirow{2}{*}{\#}&\multirow{2}{*}{Method}& \multirow{2}{*}{Train} & \multicolumn{6}{c}{Val seen} & \multicolumn{6}{c}{Val unseen} \\

    \cmidrule(r){4-9}
    \cmidrule(r){10-15}
&  ~ & ~ & TL& NE$\downarrow$ & nDTW$\uparrow$& OSR$\uparrow$& SR$\uparrow$& SPL$\uparrow$& TL& NE$\downarrow$& nDTW$\uparrow$& OSR$\uparrow$& SR$\uparrow$&SPL$\uparrow$\\
  \midrule

\rowcolor{Cerulean!20}\multicolumn{15}{l}{{Evaluate with Hight Line-of-Sight:}}\\

% & seq2seq& H& 9.37& 7.02& 0.54& 0.46& 0.33& 0.31& 9.32& 7.77& 0.47& 0.47& 0.25& 0.22\\ 

% & \multirow{3}{*}{CMA(mono)} & H& 9.26& 7.12& 0.54& 0.46& 0.37& 0.35& 8.64& 7.37& 0.51& 0.40& 0.32& 0.30\\
1 &CMA (mono)~\cite{krantz2020beyond} & - & 9.26& 7.12& 0.54& 0.46& 0.37& 0.35& 8.64& 7.37& 0.51& 0.40& 0.32& 0.30\\
% & CMA(pano) & H& 11.47& 5.20& 0.61& 0.61& 0.51& 0.45& 10.9& 6.2& 0.55& 0.52& 0.41& 0.36\\ 
2 &RecurrentBert~\cite{hong2022bridging}& - & 12.50 & 5.02& 0.58 & 0.59& 0.50& 0.44& 12.23& 5.74& 0.54& 0.53& 0.44& 0.39\\
3 & BEVBert~\cite{an2022bevbert}& - & 12.35 & 3.22 & 0.70& 0.78& 0.71& 0.63& -& 4.70& -& 0.67& 0.59& 0.50\\ 
4 &ETPNav~\cite{an2023etpnav} & - & 11.78& 3.95& -& 0.72& 0.66& 0.59& 11.99& 4.71& -& 0.65& 0.57& 0.49\\ 

\midrule

\rowcolor{Cerulean!20}\multicolumn{15}{l}{{Evaluate with Low Line-of-Sight:}}\\
% & \multirow{2}{*}{seq2seq}& \xmark & 9.15& 9.58& 0.30& 0.17& 0.06& 0.04& 9.23& 9.35& 0.30& 0.17& 0.05&0.04\\
% & & \cmark& 7.96 & 6.82 & 0.53 & 0.39 & 0.21 & 0.2 & 7.66 & 7.11 & 0.51 & 0.34 & 0.19 & 0.18\\
5 & \multirow{2}{*}{CMA (mono)~\cite{krantz2020beyond}}& \xmark & 10.29& 7.69& 0.45& 0.36& 0.23& 0.20& 7.00 & 7.75& 0.43& 0.32& 0.19&0.16\\
6 & & \cmark & 7.99& 6.43& 0.53& 0.41& 0.24& 0.21& 7.16 & 7.41 & 0.53& 0.36 & 0.22 & 0.18\\
% \midrule
% & \multirow{2}{*}{CMA(pano)}& \xmark & 12.49& 8.43& 0.31& 0.33& 0.19& 0.13& 15.05& 8.41& 0.31& 0.33& 0.18&0.13\\
% & & \cmark & 13.24 & 5.99  & 0.53 & 0.58 & 0.47 & 0.37 & 15.53 & 6.77 & 0.44 & 0.44& 0.31 & 0.23\\ 

7 &\multirow{2}{*}{RecurrentBert~\cite{hong2022bridging}} & \xmark & 16.38& 7.41& 0.37& 0.37& 0.26& 0.19& 16.41& 7.32& 0.35& 0.37& 0.23&0.16\\ 
8 & & \cmark & 14.13& 5.31& 0.54& 0.57& 0.45& 0.37& 14.45& 5.99& 0.48& 0.48& 0.37&0.26\\
9 &\multirow{2}{*}{BEVBert~\cite{an2022bevbert}}& \xmark & 22.77 & 7.55 & 0.32 & 0.38 & 0.28 & 0.18 & 22.25 & 7.58 & 0.31 & 0.37 & 0.27 & 0.17\\ 
10& & \cmark & 14.05 & 5.02 & 0.58 & 0.61 & 0.51 & 0.43 & 15.23 & 5.61 & 0.53 & 0.57 & 0.47 & 0.38\\
11 &\multirow{2}{*}{ETPNav~\cite{an2023etpnav}}& \xmark & 22.32 & 8.53 & 0.26 & 0.32 & 0.22 & 0.13 & 21.79 & 8.14 & 0.26 & 0.32 & 0.21 & 0.11\\ 
12 & & \cmark & 12.35 & 4.60 & 0.62 & 0.68 & 0.58 & 0.50 & 12.73& 5.15& 0.57& 0.60& 0.52& 0.43\\
\midrule
 
13 & GVNav (Ours)& \cmark & 12.34 & \textbf{3.88} & \textbf{0.66} & \textbf{0.70} & \textbf{0.64} & \textbf{0.56}& 13.76 & \textbf{4.89} & \textbf{0.58} & \textbf{0.62} & \textbf{0.55} & \textbf{0.45}\\

    \bottomrule
    % \end{spacing}
    \end{tabular}
    \end{center}
    \vspace{-10pt}
\end{table*}

\begin{table*}[t]
\caption{Comparison of the contribution of Navigator and Waypoint Predictor\\
N: Navigator, WP: Waypoint Predictor; R: Re-trained, F: Freezed}
\label{tab:wp}
\vspace{-10pt}
\centering
\begin{tabular}{ccccccccccccccccc}
\toprule
 \multirow{2}{*}{Method} & \multicolumn{2}{c}{N} & \multicolumn{2}{c}{WP}  & \multicolumn{6}{c}{Val seen} & \multicolumn{6}{c}{Val unseen} \\
    \cmidrule(r){2-5}
    \cmidrule(r){6-11}
    \cmidrule(r){12-17}
  & F& R & F & R & TL& NE$\downarrow$ & nDTW$\uparrow$& OSR$\uparrow$& SR$\uparrow$& SPL$\uparrow$& TL& NE$\downarrow$& nDTW$\uparrow$& OSR$\uparrow$& SR$\uparrow$&SPL$\uparrow$\\

\midrule
% \rowcolor{Cerulean!20}\multicolumn{15}{l}{\emph{Recurrent-Bert:}}\\
\multirow{4}{*}{ETPNav~\cite{an2024etpnav}}& \cmark& & \cmark& & 22.32 & 8.53 & 0.26 & 0.33 & 0.22 & 0.13 & 21.79& 8.14& 0.26& 0.32& 0.21& 0.12\\
 
& \cmark& & & \cmark& 18.63 & 6.45 & 0.41 & 0.58 & 0.38 & 0.25 & 19.67& 6.57& 0.37& 0.56& 0.39& 0.24\\ 
& & \cmark & \cmark& & 16.39 & 7.08 & 0.39 & 0.35 & 0.30 & 0.22 & 16.61& 6.61& 0.40& 0.36& 0.32& 0.23\\ 
& &\cmark & & \cmark& 12.35 & \textbf{4.60} & \textbf{0.62} & \textbf{0.68} & \textbf{0.58} & \textbf{0.50} & 12.73 & \textbf{5.15}& \textbf{0.57}& \textbf{0.60}& \textbf{0.52}& \textbf{0.43}\\ 
 \bottomrule
\vspace{-10pt}
\end{tabular}
\vspace{-15pt}
\end{table*}

In this study, we aim to evaluate the performance of serval VLN models under varying line-of-sight perspectives, transitioning from a high line-of-sight perspective (representative of human vision) to a low line-of-sight perspective (representative of small quadruped robots). This evaluation is designed to identify the performance gap caused by viewpoint discrepancies and to assess the limitations of existing VLN models in low line-of-sight scenarios. Our implementation is based on the Habitat simulator~\cite{savva2019habitat} and uses the Matterport3D (MP3D) dataset~\cite{chang2017matterport3d}, which offers photo-realistic 3D environments with both panoramic and line-of-sight variations, effectively simulating real-world conditions. We employed a two-stage training process, with the first stage involving learning on a scaled dataset generated from the HM3D, Gibson, and MP3D datasets for waypoint prediction, and the second stage training on navigation task-specific data (R2R~\cite{krantz2020beyond}). The learning rate was set to 0.0001 with a batch size of 32. Our approach was benchmarked against multiple baselines, including Seq2Seq~\cite{krantz2020beyond}, CMA (mono)~\cite{krantz2020beyond}, BEVBert~\cite{an2022bevbert}, and ETP~\cite{an2023etpnav}. Evaluation metrics included Trajectory Length (TL), Navigation Error (NE), Overall Success Rate (OSR), Success Rate (SR), and Success weighted by Path Length (SPL), which collectively provided a comprehensive assessment of model performance. In our evaluation, a navigation attempt was considered successful if the robot reached within 3 meters of the target location. Additionally, we deployed our proposed method on a Xiaomi Cyberdog for real-world tests, comparing its performance against two monocular methods and two panoramic methods, to demonstrate its robustness in diverse environments.
\subsection{Comparison on Simulated Environments}
\noindent\textbf{Impact of Changing Line-of-Sight:} We evaluated VLN models trained exclusively on high line-of-sight data (approximately 1.7 meters, simulating a human perspective) and tested them under both high and low line-of-sight conditions. This comparison revealed a substantial performance gap between the two settings, illustrating the difficulty of applying models trained on high line-of-sight visual data to small quadruped robots. As shown in Table~\ref{tab:policy}, models M\#1 and M\#5 scored 13\% lower SR scores, respectively, for the CMA and RecurrentBert. Comparing M\#3 to M\#9 and M\#4 to M\#11, the SR scores are reduced by 32\% and 36\% for BEVBert and ETPNav, respectively. The disparity between these two perspectives, particularly in landmark recognition, depth perception, and spatial awareness, leads to navigation errors, underscoring that models trained on high line-of-sight data are not directly transferable to small robots with a pronounced downward-shifting line of sight. Especially for BEVBert(M\#9, M\#10) and ETPNav(M\#11, M\#12), as they heavily rely on depth information for spatial accessibility to predict waypoint~\cite{an2023etpnav, an2022bevbert}. To address this, we re-trained exclusively on the same model configuration using low line-of-sight visual inputs and there remains a noticeable drop in performance when compared to high line-of-sight tasks that are shown in Table~\ref{tab:policy}(M\#6, M\#8, M\#10 and M\#12). This highlights the inherent limitations of re-training the model on low-perspective data without introducing additional architectural adjustments or compensatory mechanisms. The reduced performance underscores that the discrepancy between viewpoints introduces a domain gap, which cannot be bridged solely through data re-training. This finding aligns with the results discussed in Section III, Part B, which highlighted the mismatch between the oracle and local observations due to occlusions. This reflects a critical limitation in the design of existing VLN datasets, which primarily focus on high-level, human-like visual data, leaving a performance gap when applied to low perspectives. In M\#13 we compared our GVNav with current state-of-the-art methods on the R2R-CE dataset. The results demonstrate that our model outperforms the existing models on all splits in terms of NE: 0.26-0.72, nDTW: 1\%-5\%, OSR: 2\%-5\%, SR: 3\%-8\%, and SPL: 2\%-7\%. 

\begin{table*}[t]
\vspace{-5pt}
\caption{Comparison between our approach with other methods deployed on Xiaomi Cyberdog in 4 environments.}
\label{tab:tab3}
\vspace{-8pt}
\centering
\resizebox{\linewidth}{!}{
\begin{tabular}{ccccccccccccccccc}
% \hline
%  \multicolumn{17}{c}{Real Environments}\\
% \hline
\toprule
\multirow{2}{*}{Method} & \multicolumn{4}{c}{Gaming Room} & \multicolumn{4}{c}{Kitchen} &  \multicolumn{4}{c}{Lab} & \multicolumn{4}{c}{Office Area}\\
    \cmidrule(r){2-5}
    \cmidrule(r){6-9}
    \cmidrule(r){10-13}
    \cmidrule(r){14-17}
    
& TL& NE$\downarrow$ & OSR$\uparrow$& SR$\uparrow$& TL& NE$\downarrow$ & OSR$\uparrow$& SR$\uparrow$& TL& NE$\downarrow$ & OSR$\uparrow$& SR$\uparrow$& TL& NE$\downarrow$ & OSR$\uparrow$& SR$\uparrow$\\
\midrule
 Seq2seq\cite{krantz2020beyond} & - & - & 0 & 0 & - & - & 0 & 0 & - & - & 0 & 0 & - & - & 0 & 0 \\
 CMA(mono)\cite{hong2022bridging} & - & - & 0 & 0 & - & - & 0 & 0 & - & - & 0 & 0 & - & - & 0 & 0 \\ 
 BEVBert\cite{an2023bevbert} & 3.32 & 5.50 & 0.08 & 0.08 & 9.88 & 4.13 & 0.24 & 0.16 & 10.46 & 5.74 & 0.16 & 0.12 & 4.30 & 4.33 & 0.20 & 0.12\\
 ETPNav\cite{an2024etpnav} & 4.76 & \textbf{3.38} & 0.28 & 0.24 & 7.78 & 3.25 & 0.32 & 0.28 & 9.98 & \textbf{4.11} & 0.24 & 0.20 & 5.60 & 3.78 & 0.32 & 0.24\\
GVNav (Ours) & 5.95 & 3.43 & \textbf{0.36} & \textbf{0.28} & 6.68 & \textbf{2.99} & \textbf{0.48} & \textbf{0.40} & 11.43 & 4.46 & \textbf{0.28} & \textbf{0.28} & 7.44 & \textbf{3.68} & \textbf{0.48} & \textbf{0.36}\\
 \bottomrule
\end{tabular}}
\vspace{-5pt}
\end{table*}

\noindent\textbf{Navigator vs. Waypoint Predictor}:
Table~\ref{tab:wp} presents ablation experiments designed to isolate the contributions of the waypoint predictor and the navigator as individual components. The experiments demonstrate that under challenging low line-of-sight conditions, the waypoint predictor has a more pronounced impact on performance compared to the navigator. We tested ETPnav in VLN-CE R2R and freeze/re-train individual components including the navigation policy and waypoint prediction networks. The results show that Re-training the waypoint predictor significantly improves the model’s generalization and navigation accuracy, from 21\% SR to 39\% SR, even when the navigator remains frozen. In contrast, re-training the navigator without updating the waypoint predictor yields only marginal performance improvements, from 21\% SR to 32\% SR as the static waypoints limit the agent's ability to navigate effectively under low-visibility conditions. To address this limitation, we constructed a larger waypoint prediction dataset, allowing for more comprehensive training of the waypoint prediction network. We use the same evaluation metric as Hong et.al~\cite{hong2022bridging}, where $\mid$$\Delta$$\mid$ measures the difference in number of target waypoints and predicted waypoints. \%Open measures the ratio of predicted waypoints that is in open space, which is the most important factor. dC and dH are the Chamfer distance and the Hausdorff distance, respectively. As shown in Table~\ref{tab:scale}, scaling up the training dataset by 22.02 times results in a 7.99\% increase in open space (1st and 4th element in Table~\ref{tab:scale}), demonstrating the effectiveness of expanding waypoint prediction training data for low line-of-sight navigation tasks.

\begin{table}[h] 
% \scriptsize
\caption{Waypoint predictor trained with scaled-up training dataset.\\
R $\longrightarrow$ Scaled RGBD; D $\longrightarrow$ Scaled Depth Only}
\vspace{-15pt}
\label{tab:scale}
\begin{center}
\resizebox{\linewidth}{!}{
\begin{tabular}{cccccccccc}
\toprule
\multirow{2}{*}{\shortstack{Height (m)}}& \multicolumn{4}{c}{MP3D Train}& \multicolumn{4}{c}{MP3D Val\_Unseen}\\
\cmidrule(r){2-5}
\cmidrule(r){6-9}
& $\mid$$\Delta$$\mid$ & \%~Open$\uparrow$ & dC$\downarrow$ & dH$\downarrow$ & $\mid$$\Delta$$\mid$ & \%Open$\uparrow$ & dC$\downarrow$ & dH$\downarrow$\\
\midrule
1.25 \cite{hong2022bridging}& 1.30 & 82.56& 1.12& 2.13& 1.40& 79.86& 1.07&2.00\\
 0.3& 1.29& 84.23& 1.10 & 2.11 & 1.36& 81.91& 1.06 &2.00 \\
 0.3(R)& 1.29& 88.70 & \textbf{1.05}& \textbf{2.01}& 1.38& 87.16& \textbf{0.99}& \textbf{1.90}\\
 0.3(D)& 1.33 & \textbf{90.02} & 1.06 & 2.04 & 1.42 & \textbf{89.90} & 1.00 & 1.92 \\
\bottomrule
\end{tabular}
}
\vspace{-20pt}
\end{center}
\end{table}

\subsection{Comparison on Real-world Environments}

% \begin{table*}[th]
% \caption{We compared our approach with other method that deployed on Xiaomi Cyberdog, they are tested in 4 environments.}
% \label{table_example}
% \centering
% \begin{tabular}{c|c|c|c|c|c|c|c|c|c|c|c|c|c|c|c|c}
% \hline
%  \multicolumn{17}{c}{Real Environments}\\
% \hline
% & \multicolumn{4}{c|}{Gaming Room} & \multicolumn{4}{c|}{Kitchen} &  \multicolumn{4}{c}{Lab} & \multicolumn{4}{|c}{Office Area}\\
% \hline
% & TL& NE$\downarrow$ & nDTW$\uparrow$& OSR$\uparrow$& SR$\uparrow$& SPL$\uparrow$& TL& NE$\downarrow$& nDTW$\uparrow$& OSR$\uparrow$& SR$\uparrow$&SPL$\uparrow$\\
% \hline
%  seq2seq\cite{krantz2020beyond} & - & - & 0 & 0 & - & - & 0 & 0 & - & - & 0 & 0 & - & - & 0 & 0 \\
% \cline{1-17}
%  CMA(mono)\cite{hong2022bridging} & - & - & 0 & 0 & - & - & 0 & 0 & - & - & 0 & 0 & - & - & 0 & 0 \\ 
% \cline{1-17}
%  BEVBert\cite{an2023bevbert} & & & & & & & & &  & & & & & & &\\
% \cline{1-17}
%  ETP(DUET)\cite{an2024etpnav} & & & & & & & & &  & & & & & & &\\
% \cline{1-17}
%  Ours & & & & & & & & &  & & & & & & &\\
% \cline{1-17}
% \end{tabular}
% \end{table*}

We deployed our GVNav approach on a Xiaomi Cyberdog to demonstrate its capability to navigate in real-world environments based on given instructions. Our pipeline allows the robot to perform low-level point navigation, effectively enabling it to navigate in unseen environments without prior mapping. The hardware was upgraded with an Intel RealSense D455 camera for more accurate depth sensing. We integrated a 360° TTL programmable gear motor to rotate the camera in precise 30° increments, capturing 12 images to form a full panoramic view. These images were fed into our navigation model for processing. All models, including CLIP, the waypoint predictor, and our navigation policy, were executed in real-time on a laptop equipped with an NVIDIA RTX 3080 Mobile GPU (16 GB VRAM).

We evaluated our method in four distinct environments: a gaming room, a kitchen, a laboratory, and an office area, with 25 unique instructions provided for each scene. As shown in Table~\ref{tab:tab3}, our approach was successfully deployed on a real robot for Vision-and-Language Navigation (VLN) tasks in real-world settings with a low line-of-sight, and the robot effectively navigated through these diverse environments. In addition to the metrics outlined in Table~\ref{tab:tab3}, our approach outperformed other methods in both simulated and real-world environments under low line-of-sight conditions.The gaming room represented a particularly challenging environment due to its cluttered layout and limited open space. In contrast, the kitchen was a smaller but relatively open area with few branching paths. The laboratory offered a more spacious and less cluttered environment, while the office area presented a highly expansive space with numerous branching paths. These diverse environments allowed for a thorough evaluation of the robustness and adaptability of our method in various spatial and navigational complexities.

\begin{figure}[t]
      \centering
      \includegraphics[width=0.95\linewidth]{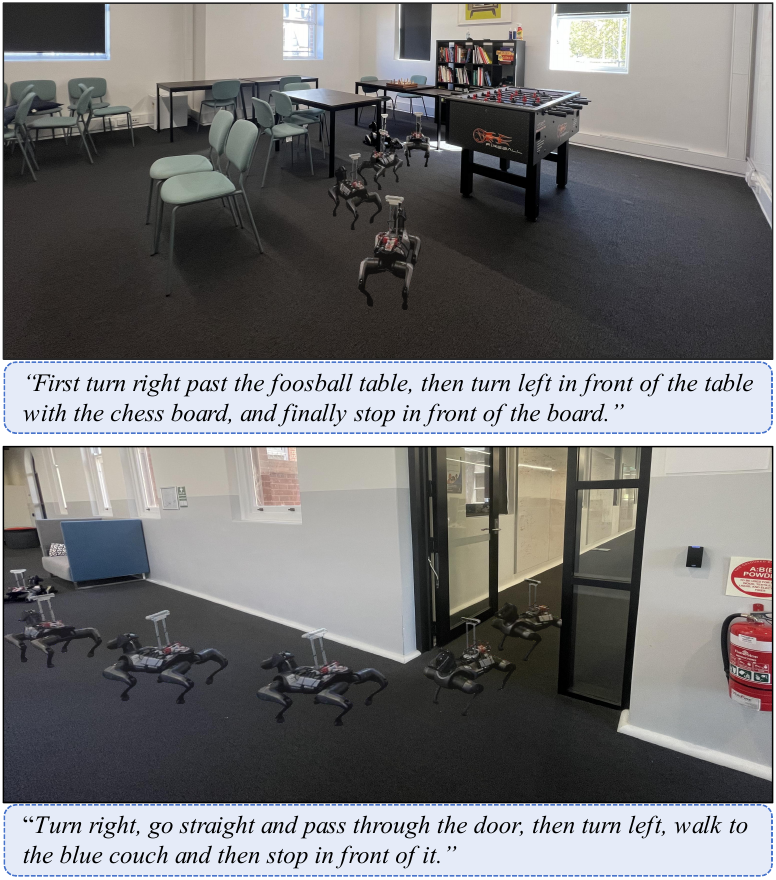}
      \vspace{-10pt}
      \caption{Real-world demo of our proposed Ground-View approach, for vision-and-language navigation. Given the human instruction, GVNav only takes 12 RGBD images as input and outputs a predicted waypoint for robotic execution.}
      \label{demo}
      \vspace{-15pt}
\end{figure}

\section{CONCLUSIONS}
In this paper, we address the key challenges of deploying VLN models on robots with low viewpoints in continuous environments. Using the Xiaomi Cyberdog as a case study, we examine the discrepancy between human commands and the robot’s limited visual input, focusing on the restricted field of view due to its low camera height. Our analysis highlights significant performance gaps caused by differences in human and robot perspectives and the limitations of monocular sensors. To address these issues, we reconstruct panoramic inputs, enhance waypoint prediction, and develop an information-gathering strategy to improve navigation performance. Our results demonstrate that bridging the visual gap between human and robot perspectives is crucial for improving the generalization and performance of VLN models.

\newpage

{
    \small
    \bibliographystyle{IEEEtran}
    \normalem 
    \bibliography{dog}

\begin{thebibliography}{10}
\providecommand{\url}[1]{#1}
\csname url@rmstyle\endcsname
\providecommand{\newblock}{\relax}
\providecommand{\bibinfo}[2]{#2}
\providecommand\BIBentrySTDinterwordspacing{\spaceskip=0pt\relax}
\providecommand\BIBentryALTinterwordstretchfactor{4}
\providecommand\BIBentryALTinterwordspacing{\spaceskip=\fontdimen2\font plus
\BIBentryALTinterwordstretchfactor\fontdimen3\font minus \fontdimen4\font\relax}
\providecommand\BIBforeignlanguage[2]{{%
\expandafter\ifx\csname l@#1\endcsname\relax
\typeout{** WARNING: IEEEtran.bst: No hyphenation pattern has been}%
\typeout{** loaded for the language `#1'. Using the pattern for}%
\typeout{** the default language instead.}%
\else
\language=\csname l@#1\endcsname
\fi
#2}}

\bibitem{anderson2018vision}
P.~Anderson, Q.~Wu, D.~Teney, J.~Bruce, M.~Johnson, N.~S{\"u}nderhauf, I.~Reid, S.~Gould, and A.~Van Den~Hengel, ``Vision-and-language navigation: Interpreting visually-grounded navigation instructions in real environments,'' in \emph{Proceedings of the IEEE conference on computer vision and pattern recognition}, 2018, pp. 3674--3683.

\bibitem{qi2020reverie}
Y.~Qi, Q.~Wu, P.~Anderson, X.~Wang, W.~Y. Wang, C.~Shen, and A.~v.~d. Hengel, ``Reverie: Remote embodied visual referring expression in real indoor environments,'' in \emph{Proceedings of the IEEE/CVF Conference on Computer Vision and Pattern Recognition}, 2020, pp. 9982--9991.

\bibitem{ku2020room}
A.~Ku, P.~Anderson, R.~Patel, E.~Ie, and J.~Baldridge, ``Room-across-room: Multilingual vision-and-language navigation with dense spatiotemporal grounding,'' in \emph{Proceedings of the 2020 Conference on Empirical Methods in Natural Language Processing (EMNLP)}, 2020, pp. 4392--4412.

\bibitem{hong2022bridging}
Y.~Hong, Z.~Wang, Q.~Wu, and S.~Gould, ``Bridging the gap between learning in discrete and continuous environments for vision-and-language navigation,'' in \emph{Proceedings of the IEEE/CVF Conference on Computer Vision and Pattern Recognition}, 2022, pp. 15\,439--15\,449.

\bibitem{wang2023scaling}
Z.~Wang, J.~Li, Y.~Hong, Y.~Wang, Q.~Wu, M.~Bansal, S.~Gould, H.~Tan, and Y.~Qiao, ``Scaling data generation in vision-and-language navigation,'' in \emph{Proceedings of the IEEE/CVF International Conference on Computer Vision}, 2023, pp. 12\,009--12\,020.

\bibitem{krantz2020beyond}
J.~Krantz, E.~Wijmans, A.~Majumdar, D.~Batra, and S.~Lee, ``Beyond the nav-graph: Vision-and-language navigation in continuous environments,'' in \emph{European Conference on Computer Vision}.\hskip 1em plus 0.5em minus 0.4em\relax Springer, 2020, pp. 104--120.

\bibitem{irshad2021hierarchical}
M.~Z. Irshad, C.-Y. Ma, and Z.~Kira, ``Hierarchical cross-modal agent for robotics vision-and-language navigation,'' in \emph{2021 IEEE International Conference on Robotics and Automation (ICRA)}.\hskip 1em plus 0.5em minus 0.4em\relax IEEE, 2021, pp. 13\,238--13\,246.

\bibitem{an2022bevbert}
D.~An, Y.~Qi, Y.~Li, Y.~Huang, L.~Wang, T.~Tan, and J.~Shao, ``Bevbert: Topo-metric map pre-training for language-guided navigation,'' \emph{arXiv preprint arXiv:2212.04385}, 2022.

\bibitem{zhang2024navid}
J.~Zhang, K.~Wang, R.~Xu, G.~Zhou, Y.~Hong, X.~Fang, Q.~Wu, Z.~Zhang, and W.~He, ``Navid: Video-based vlm plans the next step for vision-and-language navigation,'' \emph{arXiv preprint arXiv:2402.15852}, 2024.

\bibitem{wang2024sim}
Z.~Wang, X.~Li, J.~Yang, S.~Jiang, \emph{et~al.}, ``Sim-to-real transfer via 3d feature fields for vision-and-language navigation,'' \emph{arXiv preprint arXiv:2406.09798}, 2024.

\bibitem{li2024human}
M.~Li, H.~Li, Z.-Q. Cheng, Y.~Dong, Y.~Zhou, J.-Y. He, Q.~Dai, T.~Mitamura, and A.~G. Hauptmann, ``Human-aware vision-and-language navigation: Bridging simulation to reality with dynamic human interactions,'' \emph{arXiv preprint arXiv:2406.19236}, 2024.

\bibitem{yokoyama2024vlfm}
N.~Yokoyama, S.~Ha, D.~Batra, J.~Wang, and B.~Bucher, ``Vlfm: Vision-language frontier maps for zero-shot semantic navigation,'' in \emph{2024 IEEE International Conference on Robotics and Automation (ICRA)}.\hskip 1em plus 0.5em minus 0.4em\relax IEEE, 2024, pp. 42--48.

\bibitem{zhang2024vision}
Y.~Zhang, Z.~Ma, J.~Li, Y.~Qiao, Z.~Wang, J.~Chai, Q.~Wu, M.~Bansal, and P.~Kordjamshidi, ``Vision-and-language navigation today and tomorrow: A survey in the era of foundation models,'' \emph{arXiv preprint arXiv:2407.07035}, 2024.

\bibitem{wang2023gridmm}
Z.~Wang, X.~Li, J.~Yang, Y.~Liu, and S.~Jiang, ``Gridmm: Grid memory map for vision-and-language navigation,'' in \emph{Proceedings of the IEEE/CVF International Conference on Computer Vision}, 2023, pp. 15\,625--15\,636.

\bibitem{an2023etpnav}
D.~An, H.~Wang, W.~Wang, Z.~Wang, Y.~Huang, K.~He, and L.~Wang, ``Etpnav: Evolving topological planning for vision-language navigation in continuous environments,'' \emph{arXiv preprint arXiv:2304.03047}, 2023.

\bibitem{anderson2020rxr}
A.~Ku, P.~Anderson, R.~Patel, E.~Ie, and J.~Baldridge, ``Room-across-room: Multilingual vision-and-language navigation with dense spatiotemporal grounding,'' in \emph{Proceedings of the 2020 Conference on Empirical Methods in Natural Language Processing (EMNLP)}, 2020, pp. 4392--4412.

\bibitem{thomason2020cvdn}
J.~Thomason, M.~Murray, M.~Cakmak, and L.~Zettlemoyer, ``Vision-and-dialog navigation,'' in \emph{Conference on Robot Learning}, 2020, pp. 394--406.

\bibitem{fried2018speaker}
D.~Fried, R.~Hu, V.~Cirik, A.~Rohrbach, J.~Andreas, L.-P. Morency, T.~Berg-Kirkpatrick, K.~Saenko, D.~Klein, and T.~Darrell, ``Speaker-follower models for vision-and-language navigation,'' \emph{Advances in Neural Information Processing Systems}, vol.~31, 2018.

\bibitem{ma2019self}
C.-Y. Ma, J.~Lu, Z.~Wu, G.~AlRegib, Z.~Kira, R.~Socher, and C.~Xiong, ``Self-monitoring navigation agent via auxiliary progress estimation,'' \emph{arXiv preprint arXiv:1901.03035}, 2019.

\bibitem{wang2019reinforced}
X.~Wang, Q.~Huang, A.~Celikyilmaz, J.~Gao, D.~Shen, Y.-F. Wang, W.~Y. Wang, and L.~Zhang, ``Reinforced cross-modal matching and self-supervised imitation learning for vision-language navigation,'' in \emph{Proceedings of the IEEE/CVF Conference on Computer Vision and Pattern Recognition}, 2019, pp. 6629--6638.

\bibitem{tan2019envdrop}
H.~Tan, L.~Yu, and M.~Bansal, ``Learning to navigate unseen environments: Back translation with environmental dropout,'' in \emph{Proceedings of NAACL-HLT}, 2019, pp. 2610--2621.

\bibitem{ke2019tactical}
L.~Ke, X.~Li, Y.~Bisk, A.~Holtzman, Z.~Gan, J.~Liu, J.~Gao, Y.~Choi, and S.~Srinivasa, ``Tactical rewind: Self-correction via backtracking in vision-and-language navigation,'' in \emph{Proceedings of the IEEE/CVF conference on computer vision and pattern recognition}, 2019, pp. 6741--6749.

\bibitem{fu2020counterfactual}
T.-J. Fu, X.~E. Wang, M.~F. Peterson, S.~T. Grafton, M.~P. Eckstein, and W.~Y. Wang, ``Counterfactual vision-and-language navigation via adversarial path sampler,'' in \emph{European Conference on Computer Vision}.\hskip 1em plus 0.5em minus 0.4em\relax Springer, 2020, pp. 71--86.

\bibitem{qi2020object}
Y.~Qi, Z.~Pan, S.~Zhang, A.~v.~d. Hengel, and Q.~Wu, ``Object-and-action aware model for visual language navigation,'' in \emph{European Conference on Computer Vision}.\hskip 1em plus 0.5em minus 0.4em\relax Springer, 2020, pp. 303--317.

\bibitem{hong2020graph}
Y.~Hong, C.~Rodriguez, Y.~Qi, Q.~Wu, and S.~Gould, ``Language and visual entity relationship graph for agent navigation,'' \emph{Advances in Neural Information Processing Systems}, vol.~33, 2020.

\bibitem{hao2020towards}
W.~Hao, C.~Li, X.~Li, L.~Carin, and J.~Gao, ``Towards learning a generic agent for vision-and-language navigation via pre-training,'' in \emph{Proceedings of the IEEE/CVF Conference on Computer Vision and Pattern Recognition}, 2020, pp. 13\,137--13\,146.

\bibitem{li2019robust}
X.~Li, C.~Li, Q.~Xia, Y.~Bisk, A.~Celikyilmaz, J.~Gao, N.~Smith, and Y.~Choi, ``Robust navigation with language pretraining and stochastic sampling,'' \emph{arXiv preprint arXiv:1909.02244}, 2019.

\bibitem{hong2020recurrent}
Y.~Hong, Q.~Wu, Y.~Qi, C.~Rodriguez-Opazo, and S.~Gould, ``A recurrent vision-and-language bert for navigation,'' in \emph{Proceedings of the IEEE/CVF Conference on Computer Vision and Pattern Recognition (CVPR)}, June 2021, pp. 1643--1653.

\bibitem{majumdar2020improving}
A.~Majumdar, A.~Shrivastava, S.~Lee, P.~Anderson, D.~Parikh, and D.~Batra, ``Improving vision-and-language navigation with image-text pairs from the web,'' in \emph{European Conference on Computer Vision}.\hskip 1em plus 0.5em minus 0.4em\relax Springer, 2020, pp. 259--274.

\bibitem{chen2022think}
S.~Chen, P.-L. Guhur, M.~Tapaswi, C.~Schmid, and I.~Laptev, ``Think global, act local: Dual-scale graph transformer for vision-and-language navigation,'' in \emph{Proceedings of the IEEE/CVF Conference on Computer Vision and Pattern Recognition}, 2022, pp. 16\,537--16\,547.

\bibitem{anderson2018r2r}
P.~Anderson, Q.~Wu, D.~Teney, J.~Bruce, M.~Johnson, N.~S{\"u}nderhauf, I.~Reid, S.~Gould, and A.~van~den Hengel, ``Vision-and-language navigation: Interpreting visually-grounded navigation instructions in real environments,'' in \emph{Proceedings of the IEEE Conference on Computer Vision and Pattern Recognition}, 2018, pp. 3674--3683.

\bibitem{irshad2022semantically}
M.~Z. Irshad, N.~C. Mithun, Z.~Seymour, H.-P. Chiu, S.~Samarasekera, and R.~Kumar, ``Semantically-aware spatio-temporal reasoning agent for vision-and-language navigation in continuous environments,'' in \emph{2022 26th International Conference on Pattern Recognition (ICPR)}.\hskip 1em plus 0.5em minus 0.4em\relax IEEE, 2022, pp. 4065--4071.

\bibitem{raychaudhuri2021language}
S.~Raychaudhuri, S.~Wani, S.~Patel, U.~Jain, and A.~X. Chang, ``Language-aligned waypoint (law) supervision for vision-and-language navigation in continuous environments,'' \emph{arXiv preprint arXiv:2109.15207}, 2021.

\bibitem{krantz2020navgraph}
J.~Krantz, E.~Wijmans, A.~Majumdar, D.~Batra, and S.~Lee, ``Beyond the nav-graph: Vision-and-language navigation in continuous environments,'' in \emph{European Conference on Computer Vision}, 2020.

\bibitem{krantz2021waypoint}
J.~Krantz, A.~Gokaslan, D.~Batra, S.~Lee, and O.~Maksymets, ``Waypoint models for instruction-guided navigation in continuous environments,'' in \emph{Proceedings of the IEEE/CVF International Conference on Computer Vision}, 2021, pp. 15\,162--15\,171.

\bibitem{an2024etpnav}
D.~An, H.~Wang, W.~Wang, Z.~Wang, Y.~Huang, K.~He, and L.~Wang, ``Etpnav: Evolving topological planning for vision-language navigation in continuous environments,'' \emph{IEEE Transactions on Pattern Analysis and Machine Intelligence}, 2024.

\bibitem{ramakrishnan2021habitat}
S.~K. Ramakrishnan, A.~Gokaslan, E.~Wijmans, O.~Maksymets, A.~Clegg, J.~Turner, E.~Undersander, W.~Galuba, A.~Westbury, A.~X. Chang, \emph{et~al.}, ``Habitat-matterport 3d dataset (hm3d): 1000 large-scale 3d environments for embodied ai,'' \emph{arXiv preprint arXiv:2109.08238}, 2021.

\bibitem{xia2018gibson}
F.~Xia, A.~R. Zamir, Z.~He, A.~Sax, J.~Malik, and S.~Savarese, ``Gibson env: Real-world perception for embodied agents,'' in \emph{Proceedings of the IEEE Conference on Computer Vision and Pattern Recognition}, 2018, pp. 9068--9079.

\bibitem{chang2017matterport3d}
A.~Chang, A.~Dai, T.~Funkhouser, M.~Halber, M.~Niebner, M.~Savva, S.~Song, A.~Zeng, and Y.~Zhang, ``Matterport3d: Learning from rgb-d data in indoor environments,'' in \emph{2017 International Conference on 3D Vision (3DV)}.\hskip 1em plus 0.5em minus 0.4em\relax IEEE, 2017, pp. 667--676.

\bibitem{savva2019habitat}
M.~Savva, A.~Kadian, O.~Maksymets, Y.~Zhao, E.~Wijmans, B.~Jain, J.~Straub, J.~Liu, V.~Koltun, J.~Malik, \emph{et~al.}, ``Habitat: A platform for embodied ai research,'' in \emph{Proceedings of the IEEE/CVF International Conference on Computer Vision}, 2019, pp. 9339--9347.

\bibitem{an2023bevbert}
D.~An, Y.~Qi, Y.~Li, Y.~Huang, L.~Wang, T.~Tan, and J.~Shao, ``Bevbert: Multimodal map pre-training for language-guided navigation,'' in \emph{Proceedings of the IEEE/CVF International Conference on Computer Vision}, 2023, pp. 2737--2748.

\end{thebibliography}
}

\end{document}